\documentclass[a4paper,conference]{IEEEtran}
\usepackage{cite}
\ifCLASSINFOpdf
  \usepackage[pdftex]{graphicx}
  \usepackage{multirow}

\else
   \usepackage[dvips]{graphicx}
\fi
\ifCLASSOPTIONcompsoc
 \usepackage[caption=false,font=normalsize,labelfont=sf,textfont=sf]{subfig}
\else
 \usepackage[caption=false,font=footnotesize]{subfig}
\begin{document}

\title{Partitioning Image Representation in Contrastive Learning}

\author{\IEEEauthorblockN{Hyunsub Lee and Heeyoul Choi}
\IEEEauthorblockA{Department of Information and Communication Engineering\\Handong Global University, Pohang, South Korea\\
Email: \{gustjqe, heeyoul\}@gmail.com}}

\maketitle


\begin{abstract}
In contrastive learning in the image domain, the anchor and positive samples are forced to have as close representations as possible. However, forcing the two samples to have the same representation could be misleading because the data augmentation techniques make the two samples different.
In this paper, we introduce a new representation, {\it partitioned representation}, which can learn both common and unique features of the anchor and positive samples in contrastive learning. The partitioned representation consists of two parts: content part and style part. The content part represents common features of the class, and the style part represents own features of each sample, which can lead to the representation of the data augmentation method. We can achieve the partitioned representation simply by decomposing a loss function of contrastive learning into two terms on the two separate representations, respectively.
To evaluate our representation with two parts, we take two framework models: Variational AutoEncoder (VAE) and Bootstrap Your Own Latent (BYOL), to show the content and style's separability and confirm the generalization ability in classification, respectively. Based on the experiments, we show that our approach can separate two types of information in the VAE framework and outperforms the conventional BYOL in the classification and a few-shot learning task as downstream tasks.
\end{abstract}

\IEEEpeerreviewmaketitle

\section{Introduction}
Learning good image representation is an important issue to utilize it for various tasks. Recently, contrastive self-supervised learning has been regarded as a powerful and outstanding method to learn image representation \cite{oord2018representation,bachman2019learning,asano2019self,tian2020contrastive,misra2020self}. 
Many previous works have shown competitive results compared to supervised learning \cite{he2020momentum,chen2020simple,grill2020bootstrap,chen2021exploring}. Moreover, contrastive self-supervised learning could achieve a better result than supervised learning when bigger models are used with only 1\% labels \cite{chen2020big}. 
For contrastive learning, the objective is to minimize the distance between the representations of the anchor and positive samples and to maximize the distance between the representations of the anchor and negative samples. Here, the positive and negative samples are from the same and different classes, respectively.

\begin{figure}[h]
\centering
\subfloat[]{
    \includegraphics[width=0.20\textwidth]{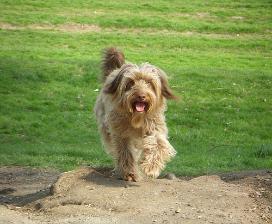}}\hfill
\subfloat[]{
    \includegraphics[width=0.17\textwidth]{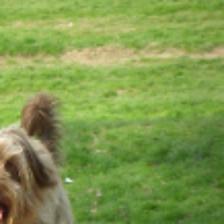} \hspace{0.1in}}
\subfloat[]{
    \includegraphics[width=0.17\textwidth]{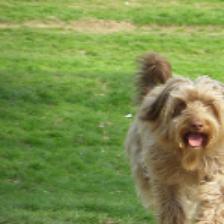}}\hfill
\subfloat[]{
    \includegraphics[width=0.17\textwidth]{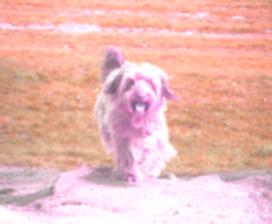}
    \hspace{0.1in}}
\subfloat[]{
    \includegraphics[width=0.17\textwidth]{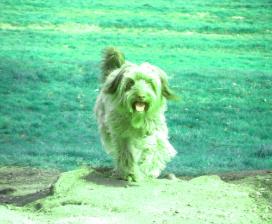}}\hfill
\subfloat[]{
    \includegraphics[width=0.17\textwidth]{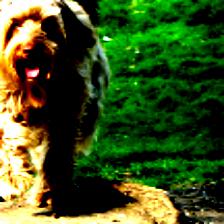}
    \hspace{0.1in}}
\subfloat[]{
    \includegraphics[width=0.17\textwidth]{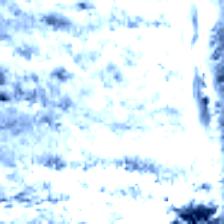}}\hfill
\caption{Different augmentation methods generate different image: (a) original image, (b-e) augmented images applying random crop, random flip, and color distortions on (a), (f-g) combination of multiple operations on (a).}
\label{fig:ResultofAugmentation}
\end{figure}

However, though contrastive learning forces the representations of the anchor and positive samples to be close to each other, they are actually different images. As shown in Figure \ref{fig:ResultofAugmentation}, the operations for multiple data augmentations can change the image drastically. Therefore, it seems unreasonable to consider the representations of the two samples to be the same, and forcing the two samples to have the exact same representation would degrade the quality of the representation.
Moreover, forcing two samples to be the same point in a single vector space could be misleading because deep neural networks tend to extract any available features to minimize the objective function on the training samples even if the features are not semantic \cite{ilyas2019adversarial,ahmed2020systematic}. Minimizing the discrepancy in a single representation could be harmful for models to learn semantic features for a given task, which could lead to a sub-optimal solution by focusing on the non-semantic features.

To overcome such issues, we introduce a new representation method, {\it partitioned representation}, which splits the representation into two parts: one for common features (or content) and the other for unique features (or style). The content part represents class-related information shared in both anchor and positive samples. Note that the anchor and positive samples belong to the same class. The style part represents unique information of each sample which is a variation within one class generated by data augmentation. 

To learn such a partitioned representation, the anchor and positive samples pull each other in the content part and push each other in the style part. This can be done by decomposing a loss function of contrastive learning for positive pairs into two terms on the two separate representations, respectively.

To evaluate the quality of our partitioned representation, we take two framework models: Variational AutoEncoder (VAE) and Bootstrap Your Own Latent (BYOL), to show the separability of the content and style features and to confirm the generalization ability in classification, respectively. 
The experiments show that we can separate two types of features in the partitioned representation in the VAE framework, and the partitioned representation outperforms the conventional BYOL with a better representation in terms of the classification accuracy in downstream tasks.

\section{Related Work}
\subsection{Contrastive Self-Supervised Learning}
Contrastive self-supervised learning is capable of learning representations using unlabelled data. In the image domain, contrastive methods have achieved state-of-the-art results on various downstream tasks. Those methods have their own design to choose the anchor and positive samples. Contrastive Predictive Coding for image (CPCv2) learns how to extract the representations of image from multiple image patches \cite{henaff2020data}. It samples anchor and positive patches from the same image, which are forced to have high mutual information. Augmented Multiscale Deep InfoMax (AMDIM) maximizes the mutual information between intermediate features of the anchor and positive samples in CNN layers \cite{bachman2019learning}. Also, Simple Framework for Contrastive Learning of Visual Representations (SimCLR) maximizes cosine similarity between header representation of the anchor and positive samples \cite{chen2020simple}.
Data augmentation generates the anchor and positive samples with two different views from the same images, and it can be combination of geometric operations (random crop and random flip) and color distortions (color jitter, color drop, solarization, etc.). 
These contrastive methods need negative samples for training which are usually from different images in the mini-batch, and the negative and positive samples can be selected more efficiently with label as in Supervised Contrastive Learning (SupCon) \cite{khosla2020supervised}.

Compared with the methods above, Bootstrap Your Own Latent (BYOL) obtains image representation without any negative sample \cite{grill2020bootstrap}. It aims to minimize only the Euclidean distance between the representations of the anchor and positive samples from two views on asymmetric network architectures. In more detail, BYOL has online and target networks consisting of the different number of head layers over an encoder. The online network with the anchor images predicts the output of the target network with the positive images. 
To train the networks, the online network is updated by the objective function, but the target network is updated by the exponential moving average of the online network. That is, the representation of the anchor sample is forced to be close to the one of the positive sample. 

In this paper, we take BYOL as a baseline to focus on the relationship between the anchor and positive samples, and modify the last representation of BYOL with the partitioned representation. Then, we mainly compare our method (BYOL with the partitioned representation) to the conventional BYOL in terms of the classification and a few-shot learning task to confirm generalization abilities.

\subsection{Dividing Representation}
To our best knowledge, the first attempt to split an embedding vector of images was proposed to exploit the embedding more efficiently in deep metric learning domain \cite{sanakoyeu2019divide}. The main idea is a kind of the `divide and conquer' approach; solving a bigger task would be more challenging than solving a set of smaller ones. They divide one problem into {\it K} sub-problems which are supposed to be separate, so each divided part of the embedding becomes in charge of solving one sub-problem. By reducing the problem complexity, their method increases the convergence speed and improves generalization. They divide the embedding dimensions and dataset into multiple parts. Then, in each iteration, only one part is updated by its own learner out of {\it K} learners. However, it does not mean that content and style features are separated in the sub-problems. Rather, it is more similar to the ensemble method.

Also, we can think of the latent space of Variational AutoEncoder (VAE) \cite{kingma2013auto} as a type of the partitioned representation. VAE is a generative model consisting of an encoder and a decoder. The encoder produces a latent vector of the input, and the decoder reconstructs the input from the latent vector. By doing so, the latent vector is considered as a representation of the input. One of the advantages is the ability to learn a disentangled representation where one dimension of the latent space represents one semantic or style feature. There have been a lot of previous works which aim to boost a disentangled latent space \cite{Higgins2017betaVAELB,kim2018disentangling,chen2018isolating,hahn2019disentangling}, but it is not guaranteed to fully disentangle the latent space without inductive biases \cite{locatello2019challenging}. 
In this paper, however, we aim for a representation where some dimensions have class-related features while the others have class-independent features.

\section{Proposed Method}
In this section, we propose a new method to train a partitioned representation in the contrastive learning framework.

We start from the fact that even though the anchor and positive samples belong to the same class, they are different. By forcing two samples to be close, a model would only focus on common features of the two samples or even non-semantic features to minimize the objective. Therefore, we partition the image representation into two parts: content and style parts. The content part is supposed to learn common features between the anchor and positive samples based on class-related information, and the style part is supposed to represent unique features of each sample focusing on non-class information. Our method considers only the relationship between the anchor and positive samples, not negative samples from different classes. As in the conventional contrastive loss, we force the representations in the content part to be close. However, the style part is trained to push the samples far apart. The objective function $L_{pr}$ between the anchor and positive samples can be defined by
 \begin{equation}
    L_{pr}(x_a, x_p) = \|(f_c(x_a), f_c(x_p))\| - \alpha * \|(f_s(x_a),f_s(x_p))\|,
    \label{eqn:PRLoss}
\end{equation}
where $x_a$ and $x_p$ are an anchor and a positive samples, $f$ is a neural network with a content part $f_c$ and a style part $f_s$, which is parametrized by trainable parameters. The hyperparameter $\alpha$ controls the weight between the content and style parts. 
We train the partitioned representation on two frameworks: VAE and BYOL \cite{kingma2013auto,grill2020bootstrap}. 

\begin{figure}[ht]
    \centering
    \hspace{-0.05in}
    \includegraphics[width=1.0\linewidth]{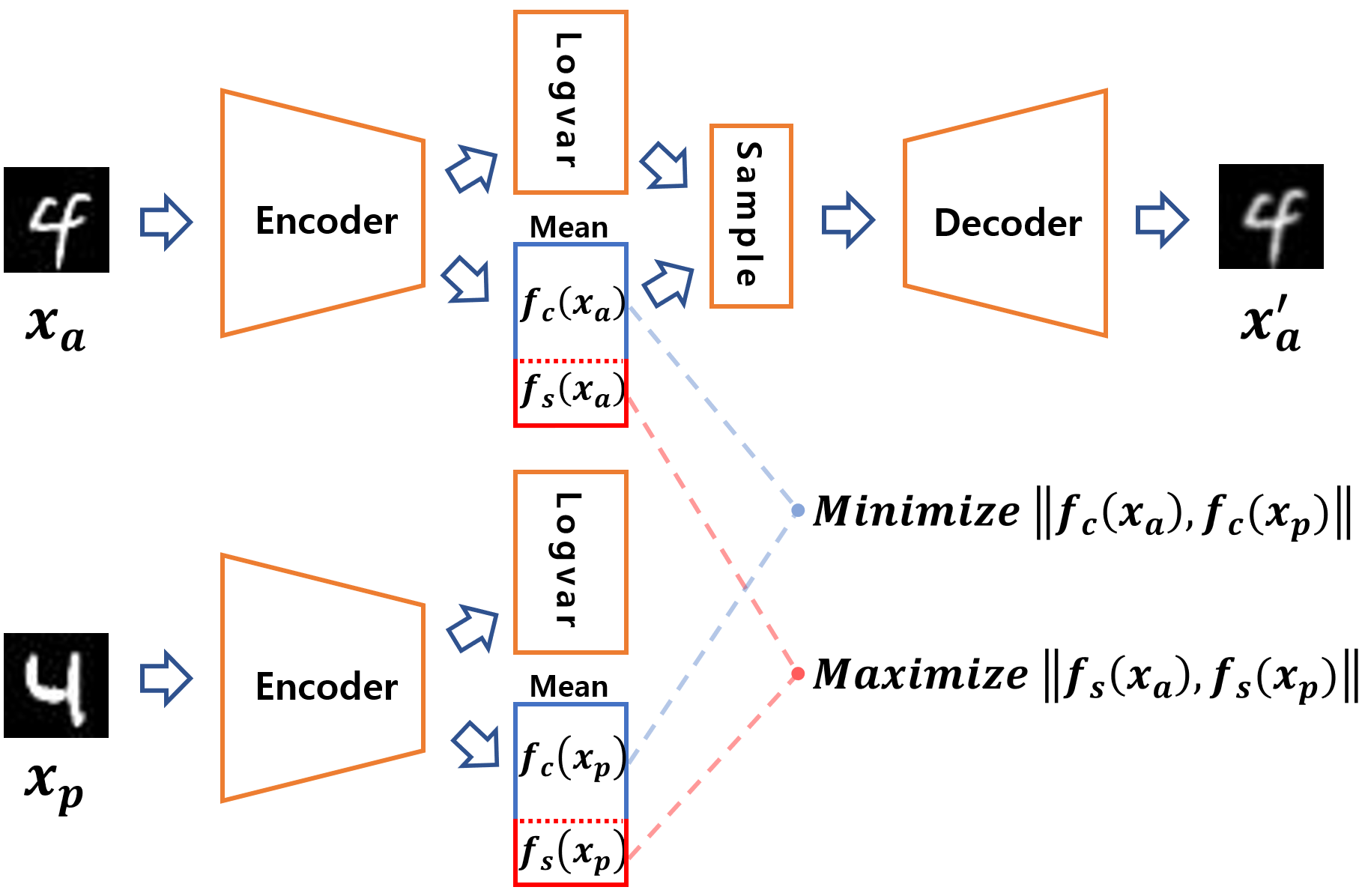}\hfill
    \caption{The overall architecture of the proposed VAE. We split the representation of the mean vector of VAE into two parts. We add the partitioned loss on ELBO as illustrated in Equation \ref{eqn:PRVAELoss}}
    \label{fig:PRVAEArchi}
\end{figure}

First, we apply our method on the VAE framework to see whether the content and style features are separated by our method. A goal of VAE is to maximize the likelihood of input $x$ by maximizing the lower bound of the likelihood called {\it Evidence Lower BOund} (ELBO). ELBO has two terms: a reconstruction error from the latent space and KL divergence between a prior distribution and data distribution obtained by the encoder as in Equation \ref{eqn:ELBO}. Without changing the two terms, we add $L_{pr}$ to $L_{ELBO}$ as in Equation \ref{eqn:PRVAELoss}, where $f$ is a mean vector obtained by the encoder. We choose a positive sample from the same class as the anchor sample as presented in Figure \ref{fig:PRVAEArchi}.
\begin{equation}
    L_{ELBO} = L_{Recons} + L_{KL}.
    \label{eqn:ELBO}
\end{equation}    
\begin{equation}
    L_{pr\_vae}(x_a, x_p) = L_{ELBO}(x_a) + L_{pr}(x_a, x_p).
    \label{eqn:PRVAELoss}
\end{equation}

\begin{figure}[ht]
    \centering
    \hspace{-0.05in}
    \includegraphics[width=1.0\linewidth]{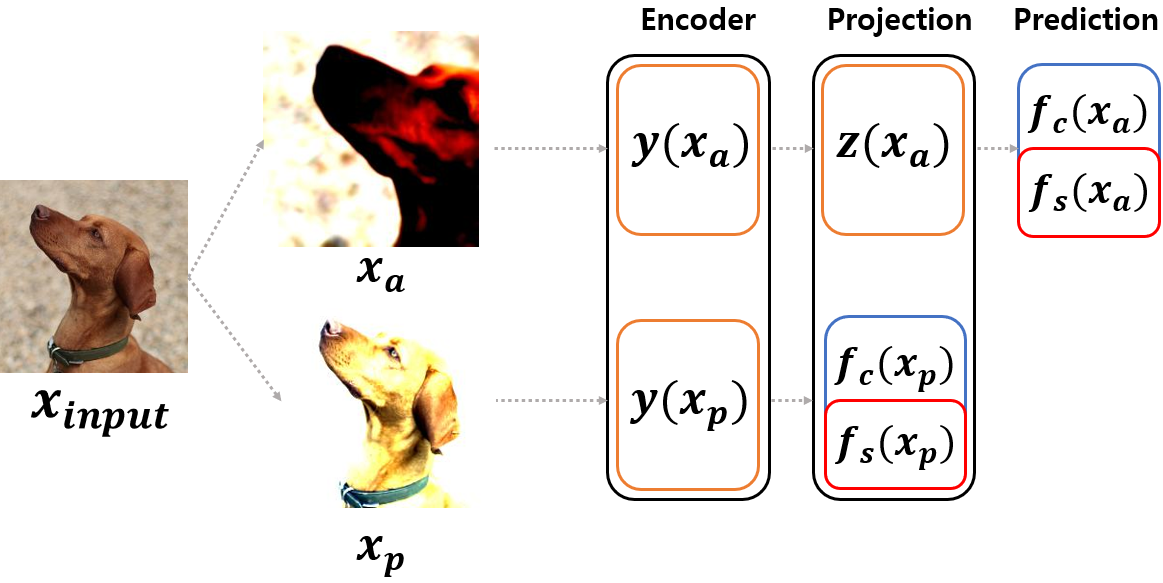}
    \caption{The proposed BYOL. We follow the architecture of the conventional BYOL except only the last representations. We divide the last representation of the online and target networks into content and style parts.}
    \label{fig:BYOLArchi}
\end{figure}

To see the generalization ability of the partitioned representation, we train the partitioned representation in the BYOL framework. BYOL minimizes the distance between the prediction vector of the anchor sample and the projection vector of the positive sample from the online and target networks. It generates two samples by applying multiple augmentations on one image, such as a random horizontal flip, random crop, Gaussian blur, and color distortions. As illustrated in Figure \ref{fig:BYOLArchi}, we directly apply Equation \ref{eqn:PRLoss} on the last representation of both networks by manually partitioning the representation into two parts instead of minimizing the entire representation.

\section{Experiment Results}
We conducted our experiments with the two frameworks: VAE and BYOL. The proposed VAE model exploits the positive sample in the supervised way while the proposed BYOL model generates the anchor and positive samples by applying the data augmentation techniques to the same image.
 
\subsection{Variational AutoEncoder}
\subsubsection{Qualitative Results}
In this section, we check qualitatively if the partitioned representation can split content and style features, especially on VAE. A traversal map of the latent space of VAE can show what features are learned by changing the value of only one dimension in the latent space. 
We used two datasets: Fashion MNIST \cite{xiao2017fashion} and colored MNIST where we add a feature about color on MNIST \cite{deng2012mnist}.

We train a VAE model with the partitioned representation and then visualize the latent space to understand what is represented in each dimension. The model has three convolution layers for the encoder and decoder, and 10 dimensions for the latent space with 7 for the content part and 3 for the style part.
For each layer, the kernel size, stride, and padding size are 3, 2, and 1, respectively. Also, the channel size of the layers in the encoder are 32, 64, and 128, respectively (reverse order for the decoder). We use the Adam optimizer with the learning rate of 0.001. 
For contrastive learning, positive images are randomly selected from the same class as the anchor image.

\begin{figure}[ht]
    \centering
    \hspace{-0.05in}
    \includegraphics[width=1.0\linewidth]{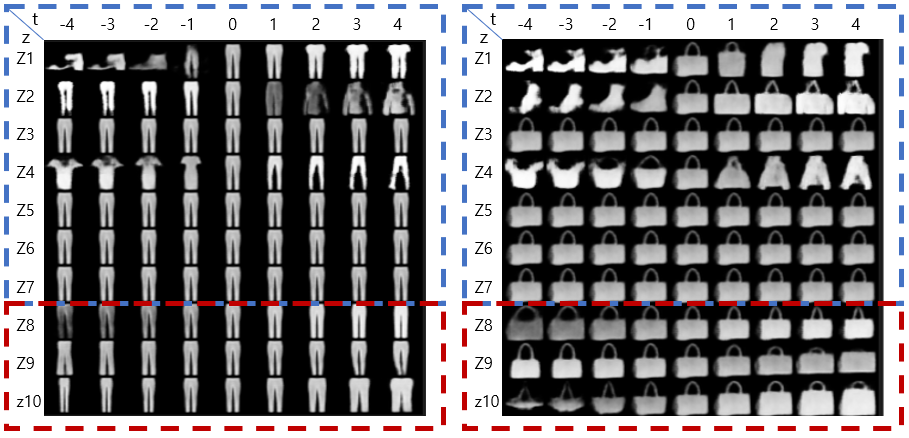}
    \caption{Traversal maps for Fashion MNIST. The first seven dimensions are for the content part (blue dotted box) and the other dimensions are for the style part (red dotted box). The content and style parts represent variations between classes and within a class, respectively.}
    \label{fig:FashionTraversal}
\end{figure}

We first train the model on the Fashion MNIST dataset, and Figure \ref{fig:FashionTraversal} shows traversal maps of the latent space. In each traversal map corresponding to one row, the center image is the reconstructed image from the latent space. Given the variance $\sigma_i$ for each dimension, we add $t\times\sigma_i$ with $t \in [-4, 4]$ on only one dimension of the latent vector and then reconstruct the images from the manipulated latent vectors. As shown in Figure \ref{fig:FashionTraversal}, the first seven dimensions (content part) show a variation from one class to another class, representing class-related features. However, the last three dimensions (style part) show variations within the class of the input image, which means they do not represent class-dependent features. Note that the common features between the anchor and positive images are class-related features because they are from the same class, otherwise, the unique features are about the style of each image.

\begin{figure}[h]
    \centering
    \hspace{-0.1in}
    \subfloat[]{
        \includegraphics[width=0.49\linewidth]{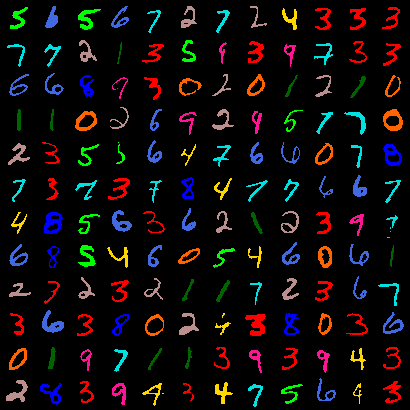}}
    \subfloat[]{
        \includegraphics[width=0.49\linewidth]{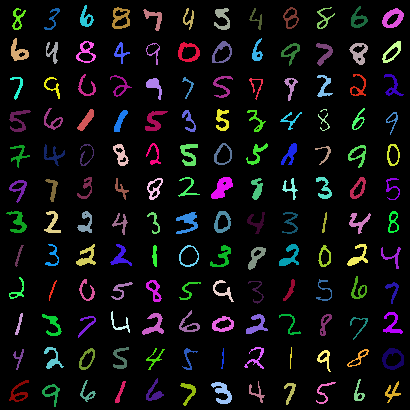}}
    \caption{Samples of color MNIST. (a) For the biased case, each digit has its own color as a class related feature. (b) For the unbiased case, every sample is colored by random color. That is, color is a non-class related feature.}
    \label{fig:ColorMNIST}
\end{figure}

For further investigation, we inject a content or style feature on purpose with colored MNIST. There are two cases of color injection: biased and unbiased cases as shown in Figure \ref{fig:ColorMNIST}. 
The biased one is the case where each digit has its unique color, which means that color is a class-related feature shared in the anchor and positive samples. The channel values for each biased color are listed in Table \ref{tab:BiasedColor}.  The unbiased one is the case where we color randomly every image, which means that color is the class-independent feature. Note that the anchor and positive samples share the digit information, not style like rotation, thickness, and font.

\begin{table}[h]
\caption{Channel Values of Biased Color for Classes.}
\centering
\begin{tabular}{c|rrr}
\hline
Class & Red & Green & Blue \\ 
\hline
0     & 255 & 100 & 0      \\
1     & 0   & 100 & 0      \\
2     & 188 & 143 & 143    \\
3     & 255 & 0   & 0      \\
4     & 255 & 215 & 0      \\
5     & 0   & 255 & 0      \\
6     & 65  & 105 & 225    \\
7     & 0   & 225 & 255    \\
8     & 0   & 0   & 255    \\
9     & 255 & 20  & 147    \\ 
\hline
\end{tabular}
\label{tab:BiasedColor}
\end{table}

\begin{figure*}[!h]\centering
    \subfloat[]{
        \includegraphics[width = 0.3\linewidth, height = 0.6\columnwidth]{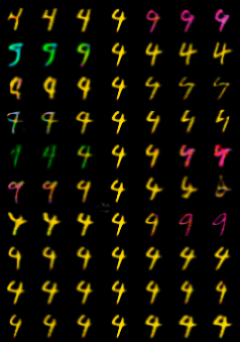}\hspace{0.1in}
        \label{fig:Biaseda}}
    \subfloat[]{
        \includegraphics[width = 0.3\linewidth, height = 0.6\columnwidth]{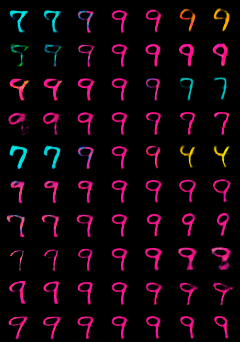}\hspace{0.1in}
        \label{fig:Biasedb}}
    \subfloat[]{
        \includegraphics[width = 0.3\linewidth, height = 0.6\columnwidth]{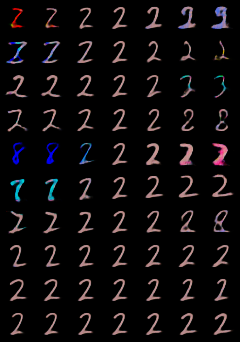}
        \label{fig:Biasedc}}
        \hfill
    \subfloat[]{
        \includegraphics[width = 0.3\linewidth, height = 0.6\columnwidth]{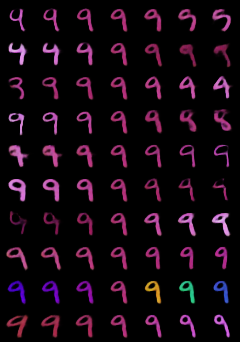}\hspace{0.1in}
        \label{fig:Unbiasedd}}
    \subfloat[]{
        \includegraphics[width = 0.3\linewidth, height = 0.6\columnwidth]{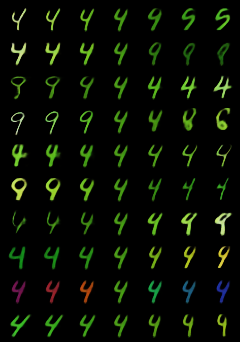}\hspace{0.1in}
        \label{fig:Unbiasede}}
    \subfloat[]{
        \includegraphics[width = 0.3\linewidth, height = 0.6\columnwidth]{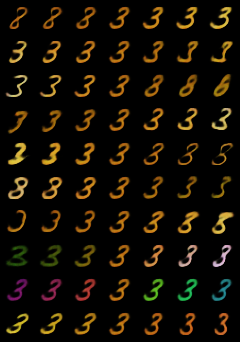}
        \label{fig:Unbiasedf}}
    \caption{Traversal maps. (a-c) In the biased color case, the color appears in first 7 dimensions which belong to the content part. The color of digits changes as the digit changes. (d-f) In the unbiased color case, the color which is not class related feature appears in the last 3 dimensions which are style part.}
    \label{fig:ColorTraversal}
\end{figure*}

As shown in Figure \ref{fig:ColorTraversal}, in the biased color case (first row), the color appears in the content part because the color is a shared feature between the anchor and positive images. We can see that color changes only when the class (digit) changes in the latent space, and also style features appear in the style part. In the unbiased color case (second row), the color appears in the style part because color is one of the prevalent variations in the dataset. As a result, the color turns up in the style part with other style features. We observe that by partitioning the representation into the two parts, our approach successfully separates two types of features in the VAE framework. 

% To show how much improvement our method achieves compared to the vanilla VAE model, we present traversal maps from the vanilla VAE model in Figure \ref{fig:VAEMaps}. The traversal maps show mixed representations of color information where the color information randomly appears in the latent space. Note that our method could force the color information at least in the style dimensions as presented in Figure \ref{fig:ColorTraversal}.

% \begin{figure}[h]  
%     \centering
%     \hspace{-0.075in}
%     \subfloat{
%         \includegraphics[width=0.47\linewidth]{Images/tarversal_conventional_VAE_1.png}}
%     \hspace{0.01in}    
%     \subfloat{
%         \includegraphics[width=0.47\linewidth]{Images/tarversal_conventional_VAE_2.png}}
% \caption{Traversal maps of the vanilla VAE model trained on unbiased color MNIST. The color randomly appears in the latent space. See Figures \ref{fig:Unbiasedd}, \ref{fig:Unbiasede}, and \ref{fig:Unbiasedf} for comparison.}
% \label{fig:VAEMaps}
% \end{figure} 

\begin{figure}[h]
    \centering
    \hspace{-0.1in}
    \subfloat[]{
        \includegraphics[width=0.40\linewidth]{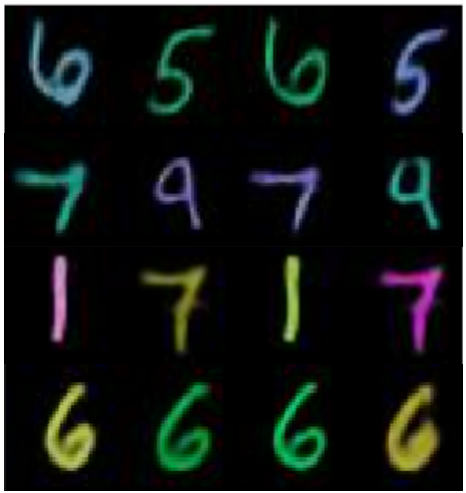}}
    \hspace{0.11in}
    \subfloat[]{
        \includegraphics[width=0.49\linewidth]{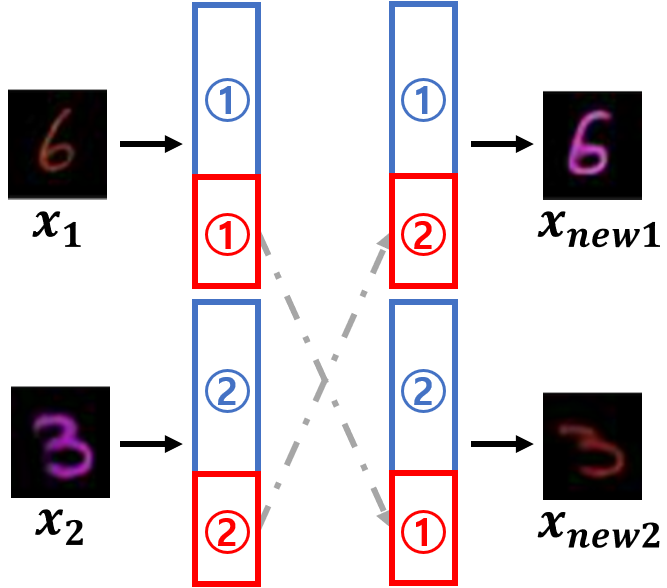}}
    \caption{Application of the partitioned representation. (a) In each row, the first two images are inputs, and the last two images are the reconstructed ones with the style of the other input image. (b) We can generate a new sample by blending the content part of one image and the style part of the other image.}
    \label{fig:VAE_Application}
\end{figure}

As an application of the partitioned representation, it is possible to generate a new sample by switching the content and style parts of two samples as in Figure \ref{fig:VAE_Application}. In Figure \ref{fig:VAE_Application}(a), the first two images of each low are the input images to VAE trained on unbiased color MNIST, and the last two images are the reconstructed ones with a switched style. Figure \ref{fig:VAE_Application}(b) shows how to obtain the new images. We exchange the style part of the mean vectors from the two input images. Then, we reconstruct the new samples from the new mean vector which consists of the content part of one image and the style part of the other image. As a result, the new samples represent the same digit with the style of the other image.

\subsubsection{Quantitative Results}
In this section, we investigate the effect of the partitioned representation in classification tasks.
We first train an additional linear classifier over the mean vector of the trained VAE model to classify the mean vector. To see the effect of the model, we add Gaussian noise at the style part or the content part of the latent variable.
%on the biased color and the unbiased color MNIST datasets, 
The test cases can be summarized as follows:

\begin{itemize}
\setlength\itemsep{0.4em}
    % \item \textbf{Case 1:} The color of the input image is changed randomly.
    %Coloring the digits with random colors.
    
    % \item \textbf{Case 2:} The color of the input image is replaced with one of the 10 biased colors as in Table \ref{tab:BiasedColor}. 
    %Coloring the digits with a different biased color from its predetermined biased color.
    
    \item \textbf{Noise on Style:} Gaussian noise is added to the style part of the mean vector. 
    %Adding Gaussian noise to the style part of the mean vector.
    
    \item \textbf{Noise on Content:} Gaussian noise is added to the content part of the mean vector. 
    %Adding Gaussian noise to the content part of the mean vector.
\end{itemize}

By adding noise, we perturb the style or content part of the mean vector to check if the proposed method can split content and style features.

Table \ref{tab:QuantiVAE} shows the results with both datasets. For the VAE model trained on the biased color dataset, the classifier achieves 99\% on the test set from the biased color dataset. When Gaussian noise is added, we observe that the style part is more robust to the noise than the content part, because the style features have nothing to do with the classification task while content features are class-related. Similarly, for the unbiased color dataset, adding noise to the style part is more robust than adding noise to the content part.
Another interesting point is that the biased color case degrades less than the unbiased color case when the noise is added to the content part. Since the biased color case has a strong class-related feature with color in the content part, the representation can be more robust against random noise on it.

\begin{table}[h]
\centering
\caption{Classification Accuracy (\%) of a Linear Classifier on the Mean Vector of VAE with Noise.}
\begin{tabular}{c|c|c|c}
\hline
Dataset & Test Set & Noise on Style  & Noise on Content  \\ \hline
Biased Color & 99.01 & 98.18 & 72.36 \\
Unbiased Color & 91.55 & 90.32 & 43.54 \\ \hline
\end{tabular}
\label{tab:QuantiVAE}
\end{table}

Additionally, Table \ref{tab:QuantiVAEforNoise} shows how accuracy changes as the intensity of the noise increases. Given the Gaussian noise $n \sim N(0,1)$ and the level of intensity $t \in [1, 4]$, we add $t \times n$ to the content or style part. For both color datasets, the style part is relatively more robust than the content part against the noise especially when the noise intensity increases.

\begin{table}[h]
\centering
\caption{Classification Accuracy (\%) of a Linear Classifier with Varying Noise Intensity.}
\begin{tabular}{c|c|c|c|c|c}
\hline
Dataset & Noise on  & t = 1 & t = 2 & t = 3 & t = 4 \\ \hline
\multirow{2}{*}{Biased Color}   & Style   & 98.2 & 94.2 & 86.4 & 78.0 \\
& Content & 72.4 & 41.5 & 29.0 & 24.1 \\ \hline
\multirow{2}{*}{Unbiased Color} & Style   & 90.3 & 86.2 & 80.9 & 74.7 \\
& Content & 43.5 & 25.4 & 19.7 & 16.7 \\ \hline
\end{tabular}
\label{tab:QuantiVAEforNoise}
\end{table} 

\subsection{BYOL}

\subsubsection{Quantitative Result}
In this section, we evaluate the generalization ability of the partitioned representation in various tasks. In all experiments, we compare our approach (i.e., BYOL with the partitioned representation) to the conventional BYOL in terms of the accuracy of downstream tasks. For the partitioned representation, we divide 256 dimensions of representation of the conventional BYOL: 192 dimensions for the content part and 64 dimensions for the style part. Our implementation is based on the official implementation given by \cite{grill2020bootstrap}.

First, we train the two BYOL models based on the Resnet18 encoder on the STL10 dataset, which consists of 100k unlabeled images and 13k labeled images for ten classes (5k for training and 8k for the test, respectively) \cite{pmlr-v15-coates11a}. We follow the conventional BYOL for the implementation of data augmentation and training strategy \cite{grill2020bootstrap}. We also follow the linear evaluation procedures described in \cite{chen2020simple} to measure the linear separability of representations. We first pretrain both BYOLs on unlabeled data over 800 epochs and then train a linear classifier on 5k training data over a frozen encoder. Finally, we measure the classification performance of the test data for the linear evaluation task. As a result, our method achieves the accuracy of $79.7\%$, which is $1.4\%p$ higher than the conventional BYOL.

% \begin{figure}[!h]
% \centering
% \hspace{-0.05in}
% \includegraphics[width=1.0\linewidth]{Images/training-curve.png}
% \caption{The online classification accuracy during 40 epochs on ImageNet with the conventional BYOL and ours.}
% \label{fig:BYOLCurve}
% \end{figure}

We also conducted the same experiment on the BYOL models over 40 epochs with 256 batch sizes on ImageNet \cite{deng2009imagenet}, which consists of 1.28M training images and 50k validation images.
% Figure \ref{fig:BYOLCurve} shows that our method outperforms the conventional BYOL in terms of an online classification during training. The online classification is conducted by training a linear layer to classify mini-batch samples during training, following the implementation in \cite{grill2020bootstrap}. 
As presented in Table \ref{tab:LinearEvalResult} for the linear evaluation task, our method gets $1.0\%p$ greater accuracy than the conventional BYOL. 
We think that a model becomes flexible to extract more effective features with the partitioned representation which leads to better generalization. 
% The classification results after training for 40 epochs are presented in Table \ref{tab:LinearEvalResult}.

\begin{table}[h]
\centering
\caption{Linear Evaluation of the Conventional BYOL and Ours.}
%\resizebox{0.22\textwidth}{!}{
\begin{tabular}{c|c|cc}
\hline
\multicolumn{1}{c|}{\multirow{2}{*}{Dataset}} & \multirow{2}{*}{Encoder} & \multicolumn{2}{c}{Accuracy(\%)} \\ \cline{3-4} 
\multicolumn{1}{c|}{}   &     & \multicolumn{1}{c|}{BYOL} & Ours \\ 
\hline
STL10     & Resnet18   & 78.3    & 79.7 \\
ImageNet  & Resnet50   & 56.0    & 57.0 \\ 
\hline
\end{tabular}%}
\label{tab:LinearEvalResult}
\end{table}

Lastly, we experiment a few-shot learning task as a downstream task of a pretrained encoder to understand the transferability of the partitioned representation, because few-shot learning from the pretrained encoder can measure the generalization ability of the pretrained encoder \cite{chen2021self}.
Few-shot learning combined with meta-learning classifies images with a few training samples, and the test images are from unseen classes during training.
It typically performs `$K$-shot $N$-way' tasks where we randomly take $N$ classes among entire classes and draw $K$ samples for each class. In every iteration, we train an encoder with $N \times K$ samples, and then test the model on unseen classes. 

It was proposed to pretrain an embedding network, AMDIM \cite{bachman2019learning}, in a self-supervised way which is supposed to be fine-tuned for a few-shot classification task \cite{chen2021self}. It has two training phases: self-supervised pretraining and meta fine-tuning. 
We replaced the embedding network of \cite{chen2021self} with the conventional BYOL or our modified BYOL.

We use the MiniImagetNet dataset \cite{vinyals2016matching}, which contains 60k images from 100 classes. Also, we follow the data composition as the baseline to divide 100 classes into 64 classes for training, 16 classes for validation, 20 classes for the test \cite{chen2021self}. First, an embedding network is pretrained on 100 classes, and meta-learning is applied to fine-tune the network as described in \cite{chen2021self}. We follow the same training process on the conventional BYOL and BYOL based on Resnet 50 with the partitioned representation on `1-shot 5-way' and `5-shot 5-way' classification tasks.

As shown in Table \ref{tab:FewShotResult}, the conventional BYOL outperforms the baseline on both tasks because the conventional BYOL has better transferability than AMDIM. We find that our method improves further. The partitioned representation extracts precise class-related features and class-unrelated features at the same time. That is, the partitioned representation disentangles the information for classification, which makes classification more efficient.

\begin{table}[h]
\centering
\hspace{-0.07in}
\caption{Few-Shot Classification Results in 95\% Confidence Interval on MiniImageNet. (`$*$' denotes the improvement compared to \cite{chen2021self}).}
\resizebox{0.485\textwidth}{!}{
\begin{tabular}{c|cc}
\hline
\multirow{2}{*}{Task} & \multicolumn{2}{c}{Accuracy(\%)} \\ 
\cline{2-3}           & \multicolumn{1}{c|}{BYOL} & Ours \\ 
\hline
1-shot 5-way    & $65.69\pm0.72$($1.66\uparrow*$) & $66.84\pm0.74$($2.81\uparrow*$)\\ 
5-shot 5-way    & $83.70\pm0.42$($2.55\uparrow*$) & $85.92\pm0.40$($4.77\uparrow*$) \\ \hline 
\end{tabular}}
\label{tab:FewShotResult}
\end{table}

\section{Conclusion}
We introduced a new representation method, {\it partitioned representation} and trained it successfully in contrastive learning where there are only the anchor and positive samples. We obtained the partitioned representation by simply dividing a single representation into the two parts: content and style parts. The proposed partitioned representation can represent common and unique features of the anchor and positive samples at the same time. We showed that our method could split two types of features with the latent space of VAE, and that the partitioned representation would result in better generalized and transferable representation with BYOL.

In this paper, we have focused on the relationship between only the anchor and positive samples. However, for the future works, the method can be extended to have negative samples as well as the anchor and positive samples.

\section*{Acknowledgement}
This research was supported by Basic Science Research Program through the National Research Foundation of Korea funded by the Ministry of Education (NRF-2022R1A2C1012633), and by Institute for Information \& communications Technology Promotion (IITP) grant funded by the Korea government(MSIT) (No. 2018-0-00749, Development of virtual network management technology based on artificial intelligence).
\bibliographystyle{IEEEtran}
\bibliography{references}

\begin{thebibliography}{10}
\providecommand{\url}[1]{#1}
\csname url@samestyle\endcsname
\providecommand{\newblock}{\relax}
\providecommand{\bibinfo}[2]{#2}
\providecommand{\BIBentrySTDinterwordspacing}{\spaceskip=0pt\relax}
\providecommand{\BIBentryALTinterwordstretchfactor}{4}
\providecommand{\BIBentryALTinterwordspacing}{\spaceskip=\fontdimen2\font plus
\BIBentryALTinterwordstretchfactor\fontdimen3\font minus
  \fontdimen4\font\relax}
\providecommand{\BIBforeignlanguage}[2]{{%
\expandafter\ifx\csname l@#1\endcsname\relax
\typeout{** WARNING: IEEEtran.bst: No hyphenation pattern has been}%
\typeout{** loaded for the language `#1'. Using the pattern for}%
\typeout{** the default language instead.}%
\else
\language=\csname l@#1\endcsname
\fi
#2}}
\providecommand{\BIBdecl}{\relax}
\BIBdecl

\bibitem{oord2018representation}
A.~v.~d. Oord, Y.~Li, and O.~Vinyals, ``Representation learning with
  contrastive predictive coding,'' \emph{arXiv preprint arXiv:1807.03748},
  2018.

\bibitem{bachman2019learning}
P.~Bachman, R.~D. Hjelm, and W.~Buchwalter, ``Learning representations by
  maximizing mutual information across views,'' \emph{arXiv preprint
  arXiv:1906.00910}, 2019.

\bibitem{asano2019self}
Y.~M. Asano, C.~Rupprecht, and A.~Vedaldi, ``Self-labelling via simultaneous
  clustering and representation learning,'' \emph{arXiv preprint
  arXiv:1911.05371}, 2019.

\bibitem{tian2020contrastive}
Y.~Tian, D.~Krishnan, and P.~Isola, ``Contrastive multiview coding,'' in
  \emph{Computer Vision--ECCV 2020: 16th European Conference, Glasgow, UK,
  August 23--28, 2020, Proceedings, Part XI 16}.\hskip 1em plus 0.5em minus
  0.4em\relax Springer, 2020, pp. 776--794.

\bibitem{misra2020self}
I.~Misra and L.~v.~d. Maaten, ``Self-supervised learning of pretext-invariant
  representations,'' in \emph{Proceedings of the IEEE/CVF Conference on
  Computer Vision and Pattern Recognition}, 2020, pp. 6707--6717.

\bibitem{he2020momentum}
K.~He, H.~Fan, Y.~Wu, S.~Xie, and R.~Girshick, ``Momentum contrast for
  unsupervised visual representation learning,'' in \emph{Proceedings of the
  IEEE/CVF Conference on Computer Vision and Pattern Recognition}, 2020, pp.
  9729--9738.

\bibitem{chen2020simple}
T.~Chen, S.~Kornblith, M.~Norouzi, and G.~Hinton, ``A simple framework for
  contrastive learning of visual representations,'' in \emph{International
  conference on machine learning}.\hskip 1em plus 0.5em minus 0.4em\relax PMLR,
  2020, pp. 1597--1607.

\bibitem{grill2020bootstrap}
J.-B. Grill, F.~Strub, F.~Altch'e, C.~Tallec, P.~H. Richemond, E.~Buchatskaya,
  C.~Doersch, B.~{\'A}. Pires, Z.~D. Guo, M.~G. Azar, B.~Piot, K.~Kavukcuoglu,
  R.~Munos, and M.~Valko, ``Bootstrap your own latent: A new approach to
  self-supervised learning,'' \emph{ArXiv}, vol. abs/2006.07733, 2020.

\bibitem{chen2021exploring}
X.~Chen and K.~He, ``Exploring simple siamese representation learning,'' in
  \emph{Proceedings of the IEEE/CVF Conference on Computer Vision and Pattern
  Recognition}, 2021, pp. 15\,750--15\,758.

\bibitem{chen2020big}
T.~Chen, S.~Kornblith, K.~Swersky, M.~Norouzi, and G.~Hinton, ``Big
  self-supervised models are strong semi-supervised learners,'' \emph{arXiv
  preprint arXiv:2006.10029}, 2020.

\bibitem{ilyas2019adversarial}
A.~Ilyas, S.~Santurkar, D.~Tsipras, L.~Engstrom, B.~Tran, and A.~Madry,
  ``Adversarial examples are not bugs, they are features,'' \emph{arXiv
  preprint arXiv:1905.02175}, 2019.

\bibitem{ahmed2020systematic}
F.~Ahmed, Y.~Bengio, H.~van Seijen, and A.~Courville, ``Systematic
  generalisation with group invariant predictions,'' in \emph{International
  Conference on Learning Representations}, 2020.

\bibitem{henaff2020data}
O.~Henaff, ``Data-efficient image recognition with contrastive predictive
  coding,'' in \emph{International Conference on Machine Learning}.\hskip 1em
  plus 0.5em minus 0.4em\relax PMLR, 2020, pp. 4182--4192.

\bibitem{khosla2020supervised}
P.~Khosla, P.~Teterwak, C.~Wang, A.~Sarna, Y.~Tian, P.~Isola, A.~Maschinot,
  C.~Liu, and D.~Krishnan, ``Supervised contrastive learning,'' \emph{Advances
  in Neural Information Processing Systems}, vol.~33, pp. 18\,661--18\,673,
  2020.

\bibitem{sanakoyeu2019divide}
A.~Sanakoyeu, V.~Tschernezki, U.~Buchler, and B.~Ommer, ``Divide and conquer
  the embedding space for metric learning,'' in \emph{Proceedings of the
  IEEE/CVF Conference on Computer Vision and Pattern Recognition}, 2019, pp.
  471--480.

\bibitem{kingma2013auto}
D.~P. Kingma and M.~Welling, ``Auto-encoding variational bayes,'' \emph{arXiv
  preprint arXiv:1312.6114}, 2013.

\bibitem{Higgins2017betaVAELB}
I.~Higgins, L.~Matthey, A.~Pal, C.~P. Burgess, X.~Glorot, M.~M. Botvinick,
  S.~Mohamed, and A.~Lerchner, ``Beta-vae: Learning basic visual concepts with
  a constrained variational framework,'' in \emph{ICLR}, 2017.

\bibitem{kim2018disentangling}
H.~Kim and A.~Mnih, ``Disentangling by factorising,'' in \emph{International
  Conference on Machine Learning}.\hskip 1em plus 0.5em minus 0.4em\relax PMLR,
  2018, pp. 2649--2658.

\bibitem{chen2018isolating}
R.~T. Chen, X.~Li, R.~Grosse, and D.~Duvenaud, ``Isolating sources of
  disentanglement in variational autoencoders,'' \emph{arXiv preprint
  arXiv:1802.04942}, 2018.

\bibitem{hahn2019disentangling}
S.~Hahn and H.~Choi, ``Disentangling latent factors of variational auto-encoder
  with whitening,'' in \emph{International Conference on Artificial Neural
  Networks}.\hskip 1em plus 0.5em minus 0.4em\relax Springer, 2019, pp.
  590--603.

\bibitem{locatello2019challenging}
F.~Locatello, S.~Bauer, M.~Lucic, G.~Raetsch, S.~Gelly, B.~Sch{\"o}lkopf, and
  O.~Bachem, ``Challenging common assumptions in the unsupervised learning of
  disentangled representations,'' in \emph{international conference on machine
  learning}.\hskip 1em plus 0.5em minus 0.4em\relax PMLR, 2019, pp. 4114--4124.

\bibitem{xiao2017fashion}
H.~Xiao, K.~Rasul, and R.~Vollgraf, ``Fashion-mnist: a novel image dataset for
  benchmarking machine learning algorithms,'' \emph{arXiv preprint
  arXiv:1708.07747}, 2017.

\bibitem{deng2012mnist}
L.~Deng, ``The mnist database of handwritten digit images for machine learning
  research,'' \emph{IEEE Signal Processing Magazine}, vol.~29, no.~6, pp.
  141--142, 2012.

\bibitem{pmlr-v15-coates11a}
A.~Coates, A.~Ng, and H.~Lee, ``An analysis of single-layer networks in
  unsupervised feature learning,'' in \emph{Proceedings of the Fourteenth
  International Conference on Artificial Intelligence and Statistics}, 2011,
  pp. 215--223.

\bibitem{deng2009imagenet}
J.~Deng, W.~Dong, R.~Socher, L.-J. Li, K.~Li, and L.~Fei-Fei, ``Imagenet: A
  large-scale hierarchical image database,'' in \emph{2009 IEEE conference on
  computer vision and pattern recognition}.\hskip 1em plus 0.5em minus
  0.4em\relax Ieee, 2009, pp. 248--255.

\bibitem{chen2021self}
D.~Chen, Y.~Chen, Y.~Li, F.~Mao, Y.~He, and H.~Xue, ``Self-supervised learning
  for few-shot image classification,'' in \emph{ICASSP 2021-2021 IEEE
  International Conference on Acoustics, Speech and Signal Processing
  (ICASSP)}.\hskip 1em plus 0.5em minus 0.4em\relax IEEE, 2021, pp. 1745--1749.

\bibitem{vinyals2016matching}
O.~Vinyals, C.~Blundell, T.~Lillicrap, D.~Wierstra \emph{et~al.}, ``Matching
  networks for one shot learning,'' \emph{Advances in neural information
  processing systems}, vol.~29, pp. 3630--3638, 2016.

\end{thebibliography}
\end{document}